\pdfoutput=1

\documentclass[11pt]{article}

\usepackage{EMNLP2023}

\usepackage{times}
\usepackage{latexsym}

\usepackage[T1]{fontenc}

\usepackage[utf8]{inputenc}

\usepackage{microtype}

\usepackage{inconsolata}

\usepackage{tabularx}
\usepackage{booktabs}
\usepackage{multirow}
\usepackage{multicol}
\usepackage{setspace}
\usepackage{longtable}
\usepackage{tablefootnote}
\usepackage{float}
\usepackage{graphicx}
\usepackage{subfigure}
\usepackage{amsmath}
\usepackage{xcolor}
\usepackage{hyperref}
\usepackage[font=small,skip=2pt]{caption}

\title{Motion Generation from Fine-grained Textual Descriptions}

\author{
    Kunhang Li $^{1,2}$, Yansong Feng $^1$\\
    $^1$ Peking University, $^2$ The University of Tokyo\\
    \href{mailto:kunhangli@g.ecc.u-tokyo.ac.jp}{\texttt{kunhangli@g.ecc.u-tokyo.ac.jp}}, \href{mailto:fengyansong@pku.edu.cn}{\texttt{fengyansong@pku.edu.cn}}
}

\begin{document}

\maketitle
\begin{abstract}
The task of \textbf{text2motion} is to generate human motion sequences from given textual descriptions, where the model explores diverse mappings from natural language instructions to human body movements. While most existing works are confined to coarse-grained motion descriptions, e.g., ``\textit{A man squats.}", fine-grained descriptions specifying movements of relevant body parts are barely explored. Models trained with coarse-grained texts may not be able to learn mappings from fine-grained motion-related words to motion primitives, resulting in the failure to generate motions from unseen descriptions. In this paper, we build a large-scale language-motion dataset specializing in fine-grained textual descriptions, FineHumanML3D, by feeding GPT-3.5-turbo with step-by-step instructions with pseudo-code compulsory checks. Accordingly, we design a new text2motion model, FineMotionDiffuse, making full use of fine-grained textual information. Our quantitative evaluation shows that FineMotionDiffuse trained on FineHumanML3D improves FID by a large margin of 0.38, compared with competitive baselines. According to the qualitative evaluation and case study, our model outperforms MotionDiffuse in generating spatially or chronologically composite motions, by learning the implicit mappings from fine-grained descriptions to the corresponding basic motions. We release our data at \href{https://github.com/KunhangL/finemotiondiffuse}{https://github.com/KunhangL/finemotiondiffuse}.
\end{abstract}

\section{Introduction}

\begin{figure*}[htbp]
    \centering
    \includegraphics[width=0.82\linewidth]{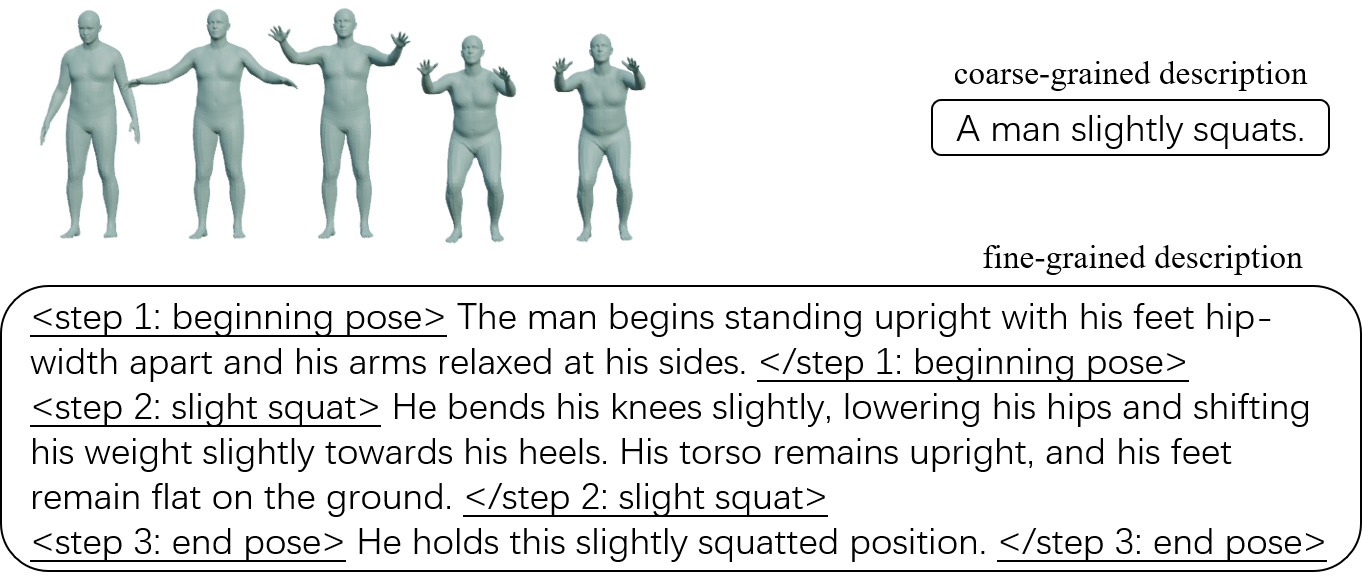}
    \caption{An example motion sequence with a coarse-grained description and its fine-grained version.}
    \label{introduction}
\end{figure*}

While most works in human motion generation from textual descriptions (\textbf{text2motion}) are devoted to designing various prediction models \citep{language2pose, Guo2022GeneratingDiverseA, Ghosh2021SynthesisOfCompostional, TM2T, TEACH, TEMOS, MotionDiffuse, MDM, FLAME, T2M-GPT}, the datasets they use only involve coarse-grained expressions, e.g., the short 4-word sentence in Figure~\ref{introduction}. However, in real-world applications, we not only describe motions in high-level coarse-grained language, but also \textbf{fine-grained instructions}. For instance, a humanoid robot needs details about movements of relevant body parts, so as to precisely perform the target motions. We find that models trained on coarse-grained descriptions cannot handle fine-grained ones. From this perspective, available datasets exhibit a bottleneck in improving the text2motion performance.

At the least, we expect the fine-grained descriptions to: 1) \textit{Be in time order}; 2) \textit{Specify spatial changes of relevant body parts}; 3) \textit{Discard unnecessary details regarding muscle tension and human feelings}; 4) \textit{Conform to human body constraints}. For the last expectation, a bad example is that the arms are raised in the previous sentences, but the next sentence assumes that the arms rest on both sides of the body.

To alleviate the data problem, a simple idea is to expand existing coarse-grained descriptions into fine-grained ones. However, it would be costly for humans to rewrite comprehensive descriptions meeting our expectations to cover a diverse range of motions. A few works attempted to utilize GPT-3 \citep{GPT3} to automatically generate fine-grained descriptions \citep{Action-GPT, SINC}. Inspired by them, we hypothesize that large language models (LLMs) learn pertinent knowledge about human bodies and physical motions, which we aim to utilize through appropriately designed prompts~\footnote{According to our pilot study, open-source LLMs like LLaMA perform worse than GPT-turbo-3.5-0301, so we decide to use this closed-source LLM.}.

For the choice of prompts used to query the LLM for fine-grained descriptions, we manually design a set of prompt templates and evaluate their performance on multiple testing texts. The final prompt instructs the LLM to generate fine-grained descriptions while summarizing them into pseudo-codes. We find that this pseudo-code compulsory check greatly enhances the stability of the LLM's generation of new descriptions. With this method, we utilize GPT-3.5-turbo to remake HumanML3D \citep{Guo2022GeneratingDiverseA} into a new language-motion dataset with fine-grained texts, FineHumanML3D. To the best of our knowledge, this is the first large-scale language-motion dataset specializing in fine-grained textual descriptions.

In Figure \ref{introduction}, we exhibit an example motion with its original coarse-grained description and the newly generated fine-grained version. Our fine-grained texts are explicitly split into steps, but the current text2motion models do not design specific mechanisms to match this property. One of the best-performing models, MotionDiffuse \citep{MotionDiffuse}, uses CLIP \citep{CLIP} with an input length limit of only 77 tokens to pre-process texts, which appears short for most of our new descriptions. Inspired by this, we propose FineMotionDiffuse, which takes fine-grained descriptions of various lengths as input, without losing any textual information due to input truncation. We also incorporate coarse-grained descriptions into FineMotionDiffuse, in order to make use of their high-level instruction-like information.

Our contributions are summarized as follows:
\begin{itemize}
    \item By applying pseudo-code compulsory checks, our final prompt elicits step-based fine-grained motion descriptions, which maintain strict chronological order while specifying movements of related body parts with appropriate granularity.
    \item By feeding our prompt into GPT-3.5-turbo, we build FineHumanML3D, the first large-scale language-motion dataset specializing in fine-grained textual descriptions.
    \item We propose FineMotionDiffuse, a new text2motion model that takes in both our fine-grained descriptions and the original coarse-grained descriptions, to take advantage of both fine textual features encoding strict step order, and coarse textual features encoding high-level instruction-like information. The quantitative evaluation shows that FineMotionDiffuse trained on FineHumanML3D improves FID by a large margin of 0.38, compared with the baseline. Moreover, human evaluation shows that FineMotionDiffuse has the smallest gap from seen basic motions to unseen composite motions (0.52 compared with 0.56 for MotionDiffuse with coarse-grained descriptions and 0.80 for MotionDiffuse with fine-grained descriptions).
\end{itemize}

\section{Related Work}

As the same motion can be described in various ways, large-scale language-motion datasets with high-quality annotations play a crucial role in solving the text2motion task. The first relevant dataset is KIT-ML \citep{KIT-ML}, but it is small in scale, with rough language descriptions and limited motion types. In comparison, HumanML3D \citep{Guo2022GeneratingDiverseA} is larger in size, with a wide variety of motion types and multiple descriptions provided for each motion. However, the language descriptions of HumanML3D still tend to be coarse-grained. While there are various composite motions, their descriptions are mostly concatenations of short verb phrases that summarize each basic motion, e.g., \textit{the figure leans down to the right, straightens, and then leans to the left}.

Action-GPT \citep{Action-GPT} uses GPT-3 \citep{GPT3} to expand the original coarse-grained texts, but they only use rough zero-shot prompts that elicit fine-grained texts with large amounts of unnecessary details. SINC \citep{SINC} mines the body parts involved in multiple motion descriptions using GPT-3, and fuses their motion sequences with such information. To generate new training data, they further synthesize the corresponding composite descriptions with conjunction words. They find the newly-trained model obtains the ability of spatial compositionality, that is, given a description involving multiple actions of different body parts, the model is able to generate the corresponding motion sequence simultaneously performing these actions.

However, the two works do not fully explore to what extent an LLM can understand, from a coarse-grained motion description, the details of time, space and human bodies. In other words, the LLM's ability to expand motion descriptions from coarse-grained to fine-grained ones still remains unknown. Moreover, we lack a large-scale language-motion dataset with high-quality fine-grained descriptions. While we expect the LLM to free humans from the laborious work of writing new annotations, it is inevitable to get imperfectly aligned fine-grained descriptions and golden motions. Therefore, we are curious about whether the LLM is able to complement enough descriptive details, so that the text2motion model trained on such noisy data still appropriately learns mappings from fine-grained motion-related descriptions to motion primitives.

For the text2motion model, the current mainstream is to map natural language expressions and motions to the same embedding space. On the one hand, works such as TM2T \citep{TM2T}, TEACH \citep{TEACH}, TEMOS \citep{TEMOS} and T2M-GPT \citep{T2M-GPT} use VAE-based architectures. On the other hand, benefiting from the recent progress of diffusion models in generation tasks, models such as MotionDiffuse \citep{MotionDiffuse}, MDM \citep{MDM} and FLAME \citep{FLAME} successfully apply such model architectures.

Of all these models, MotionDiffuse gains impressive improvements compared with previous works. It uses the multimodal model CLIP \citep{CLIP}, which fuses information from language and images by large-scale contrastive pretraining, to initialize its text embeddings. However, CLIP only enables input texts with up to 77 tokens, so the original text encoder of MotionDiffuse cannot fully encode fine-grained descriptions. We intend to overcome this problem. Furthermore, it remains unknown whether incorporation of both coarse and fine-grained texts would help improve the text2motion model's performance, which we aim to explore in this paper.

\section{Fine-grained Language-motion Dataset Construction}

We expect the LLM to properly expand motion descriptions from coarse-grained to fine-grained ones. An ideal fine-grained description is supposed to \textbf{\textit{state in chronological order}}, \textbf{\textit{specify spatial changes of relevant body parts}}, \textbf{\textit{avoid unnecessary details}} and \textbf{\textit{conform to human body constraints}}. We carefully design 8 prompts and keep the best-performing one for final dataset construction. More details about these prompts and our pilot study are in Appendix~\ref{appendix_prompts}.

\subsection{Prompt Engineering}

\begin{figure*}[htbp]
    \centering
    \includegraphics[width=\linewidth]{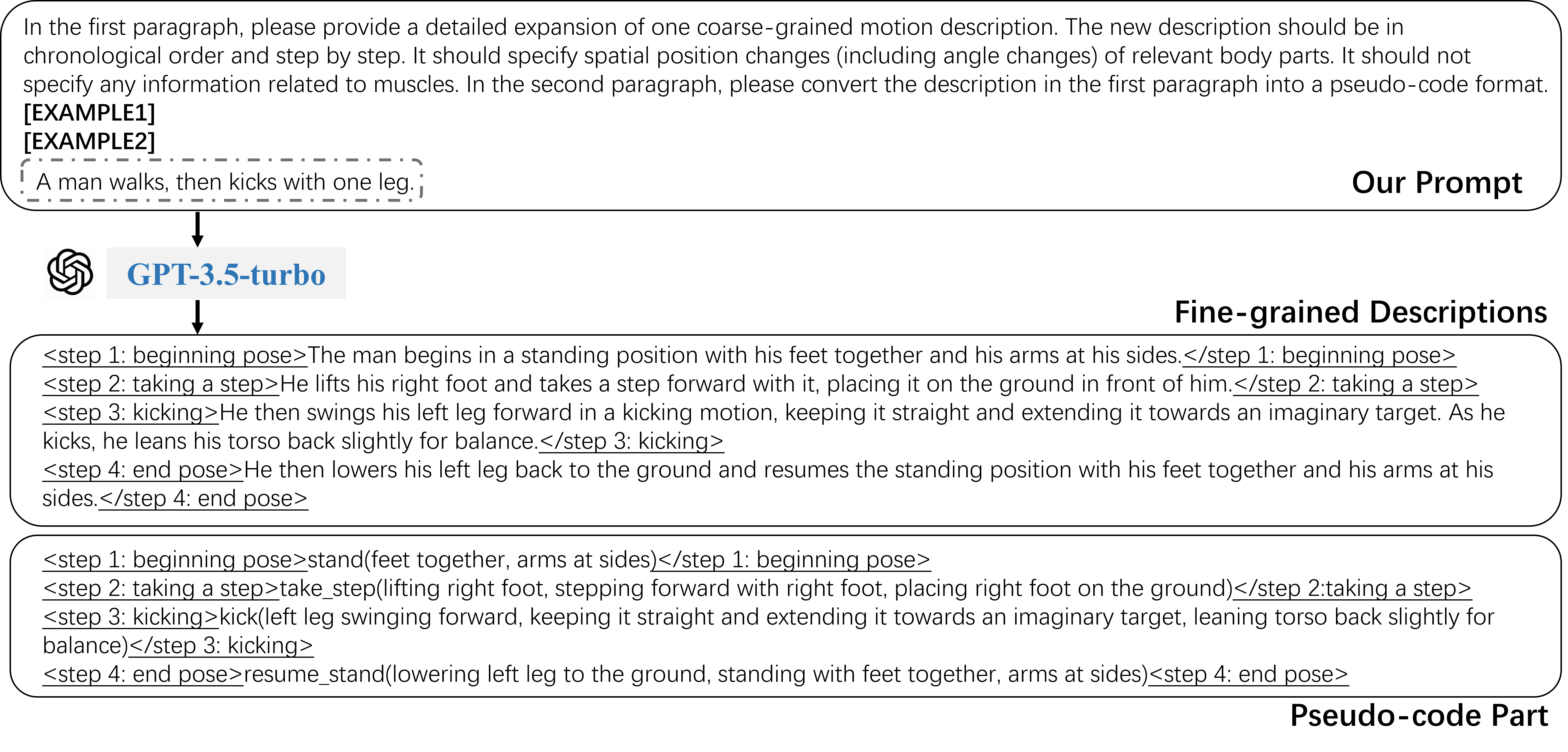}
    \caption{A fine-grained description along with its pseudo-codes acquired through \textit{P8}.}
    \label{best_prompt}
\end{figure*}

\paragraph{Pilot Prompts.} The simplest prompt, \textit{P1}, is a zero-shot prompt that asks the LLM to expand the given coarse-grained description in time order, and specify spatial position changes (including angle changes) of relevant body parts. The second prompt \textit{P2} does not require the LLM to specify anything related to muscles, and gives one instance that includes an example coarse-grained description and its fine-grained version. \textit{P3} asks the model to give a thinking process before generating the expanded descriptions. In \textit{P4}, the LLM is instead required to explicitly mark the number of steps in order, i.e., \texttt{<step n>}...\texttt{</step n>}. \textit{P4} further explicitly requires the LLM not to mention anything related to muscles.

The last prompt performs the best among the four. The descriptions it elicits are concise, without unnecessary details like muscle tension, fingertip movements, the agent's feelings, etc. Benefiting from the step marks, these descriptions are in strict time order. Surprisingly, we find the involved movements highly conform to body constraints. However, this prompt is not stable enough to assign step marks.

\paragraph{Promising Prompts.} Based on the best pilot prompt \textit{P4}, we expand step marks of \textit{P5} to specify step names in angle brackets, e.g., \texttt{<step 1:beginning pose>...</step 1:beginning pose>}. As the LLM is trained on large amounts of codes, which is believed to increase its reasoning ability \citep{Survey_GPT-3.5}, we further design \textit{P6} asking the LLM to summarize each step into a pseudo-code sequence. Prompt \textit{P7} and \textit{P8} are two-shot versions of \textit{P6}. They share the instance of \textit{P6}, differing in that the second instance of \textit{P7} is similar to the instance of \textit{P6}, while the second instance of \textit{P8} includes more code variants in pseudo-codes.

We choose 30 coarse-grained texts to test the 4 promising prompts. The main findings are that while keeping the advantages of \textit{P4}, they split steps and assign step marks more accurately. Further, we conduct a human evaluation on the quality of returned fine-grained descriptions and pseudo-codes~\footnote{In all our human evaluations, evaluators are NLP researchers with an adequate understanding of this task, and we reach an agreement on the criteria.} (Table~\ref{prompts_with_scores} in Appendix~\ref{appendix_prompts}).
Results show that the incorporation of pseudo-codes surprisingly improves the quality of returned textual descriptions. The two 2-shot prompts, \textit{P7} and \textit{P8}, are proven to perform best. But a further comparison between their worse cases shows that \textit{P7} tends to elicit pseudo-codes in formats not from the instances, a sign of uncontrollability. Therefore, we select \textit{P8} as our \textbf{final prompt}. An example input and response elicited by \textit{P8} is shown in Figure \ref{best_prompt}.

To summarize, our key findings are: 1) The LLM tends to mention muscle tension even if we do not specify this requirement, but muscles cannot be reflected in motion sequences. Therefore, we should explicitly ask the LLM not to mention anything related to muscles; 2) We can require the LLM to describe the motion step by step, and add named step marks to instances, which proves to be more useful than adding a general thinking process before expansion; 3) We can require the LLM to further summarize the fine-grained descriptions into pseudo-codes, which
increases the quality of textual descriptions; 4) We should append two diverse instances instead of one or none.

\subsection{Our FineHumanML3D Dataset}

\begin{table*}[h]
\centering
\scriptsize
\begin{tabular}{lllllllll}
\toprule[1pt]
\textbf{Dataset} & \textbf{$\#$Motions} & \textbf{$\#$Descriptions} & \textbf{$\#$Vocabulary} & \textbf{AveLen} & \textbf{$\#$Verbs} & \textbf{$\#$Nouns} & \textbf{$\#$Adpositions} & \textbf{$\#$Pronouns}\\
\midrule[1pt]
HumanML3D & 29,232 & 89,940 & 5,371 & 12.3 & 189,832 & 264,355 & 127,514 & 30,744\\
\midrule
FineHumanML3D & 29,228 & 85,646 & 8,311 & 111.7 & 1,431,013 & 2,358,655 & 1,287,520 & 1,546,554\\
\bottomrule[1pt]
\end{tabular}
\caption{\label{comparison_of_datasets}Dataset statistics for HumanML3D and our FineHumanML3D.}
\end{table*}

The original HumanML3D dataset has 44,970 coarse-grained textual descriptions for 14,616 3D human motions. The data is augmented through motion imaging and word replacement (e.g., replacing \textit{left} with \textit{right}), resulting in 89,940 descriptions for 29,232 motions.

With the selected prompt template, we expand all coarse-grained motion descriptions from HumanML3D to fine-grained ones. We delete invalid responses~\footnote{Most are sorry-like responses, e.g., \textit{I'm sorry, but the description you provided is not detailed enough...}. The rest do not conform to our predefined formats, e.g., <p> marks are used instead of named step marks.}. For each response, we discard pseudo-codes and only keep fine-grained descriptions with named step marks.

We follow~\citet{Guo2022GeneratingDiverseA} to pre-process the returned detailed descriptions, and pair each motion sequence with the fine-grained texts corresponding to its original coarse-grained ones. The final FineHumanML3D dataset has 85,646 fine-grained textual descriptions for 29,228 motions.

As shown in Table \ref{comparison_of_datasets}~\footnote{We use the tokenizer and POS tagger of spaCy 3.5 (\href{https://spacy.io/usage/v3-5}{https://spacy.io/usage/v3-5}) to calculate the statistics.}, our FineHumanML3D is rich in vocabulary, and the average description length is nearly 10 times that of HumanML3D. There are significantly more verbs, nouns and pronouns, indicating the appearance of more actions and body parts. Moreover, the large number of adpositions reflects more frequent interactions among body parts.

\subsection{Human Evaluation of FineHumanML3D}

Although \textit{P8} shows the strong ability to elicit from the LLM fine-grained descriptions with appropriate spatial details of body parts in strict chronological order, the automatically expanded texts are not guaranteed to perfectly align with the ground-truth motions. In order to assess and uncover this potential disparity, we carry out a human evaluation on a sample of FineHumanML3D, i.e., 100 randomly sampled ground-truth motions along with their fine-grained descriptions.

We classify all text-motion pairs into three categories --- zero, partial and perfect alignment, whose counts turn out to be \textbf{2 : 68 : 30}. Details of partial alignment are shown in Table~\ref{partial_alignment}~\footnote{Each text-motion pair might have several errors.}. The errors in partial alignment are categorized into inversion (e.g., left leg in the text, right leg in the motion), mismatch (partial texts not matching motion clips), redundancy (stating more than the motion performs), and deficiency (lacking descriptions of motion clips). We also show noticeable error types in redundancy and deficiency. The beginning pose type denotes the motion does not follow the beginning pose description, so for the ending pose error type. The insufficient repetition type denotes that the textual description does not repeat a single motion clip as shown by the motion, e.g., when the motion performs squatting several times, the text only describes it once.

\begin{table}[h]
\centering
\begin{tabular}{cc}
\toprule[1pt]
\textbf{Error / \#} & \textbf{Type / \#}\\
\midrule[1pt]
inversion / 26 & -\\
\midrule
mismatch / 32 & -\\
\midrule
\multirow{2}{*}{redundancy / 22} & beginning pose / 8\\
 & ending pose / 12\\
\midrule
deficiency / 11 & insufficient repetition / 5\\
\bottomrule[1pt]
\end{tabular}
\caption{\label{partial_alignment} Errors in Partial Alignment.
}
\end{table}

While only 30 out of 100 motions are perfectly aligned with the fine-grained descriptions, perfect alignment is an exceptionally high bar. In the majority of cases, partial alignment indeed captures the correct time order and relationships among core motions. Issues like redundancy or deficiency are often trivial in nature. Substantive errors rarely occur outside of very complicated motions, e.g., juggling while doing the moonwalk. Therefore, despite the partial disagreement between fine-grained descriptions and golden motions, we aim to verify the usefulness and effectiveness of our data generation method through the performance of the downstream text2motion task.

\section{Our Model}

\begin{figure*}[htbp]
    \centering
    \includegraphics[width=\linewidth]{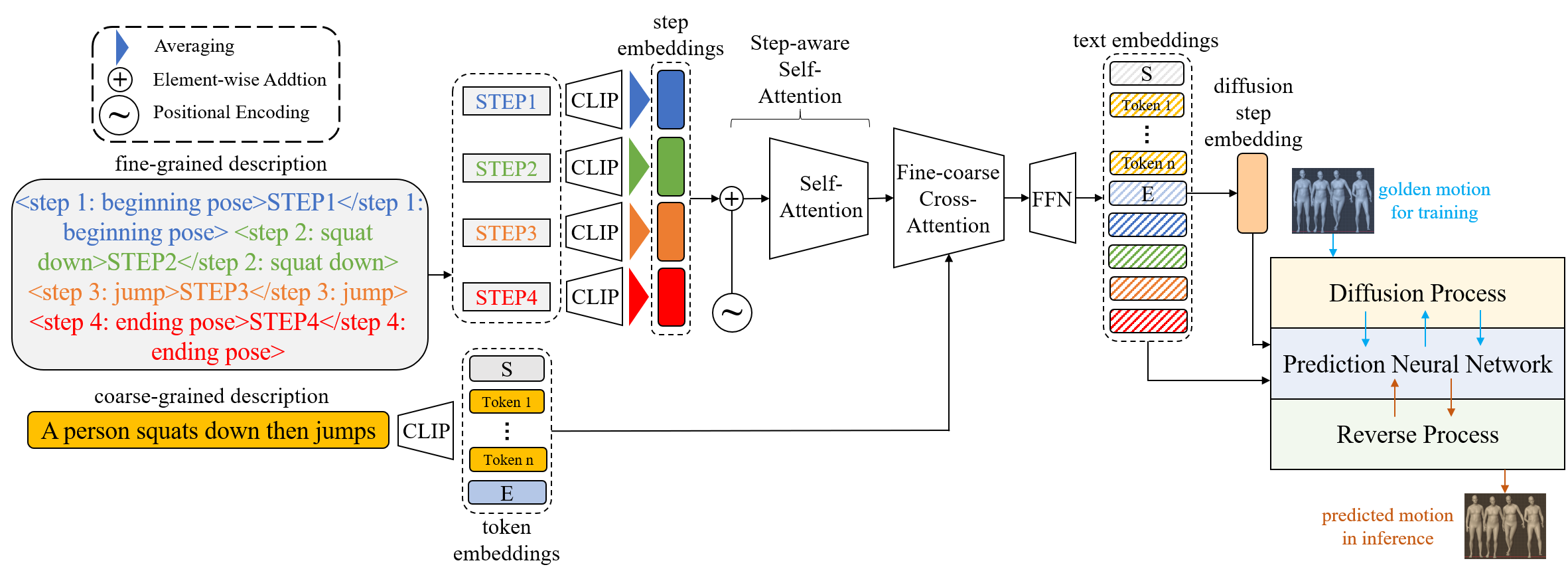}
    \caption{\textbf{An overview of our FineMotionDiffuse model.} In the diffusion block (right), blue lines indicate training flow, and brown lines for inference.}
    \label{pipeline}
\end{figure*}

We design a new text2motion model, FineMotionDiffuse, which is composed of a fine-text encoder, a step-aware self-attention block, a coarse-text encoder, a fine-coarse cross-attention block, and a diffusion block, to conduct training and inference on FineHumanML3D. An overview of our FineMotionDiffuse model is shown in Figure \ref{pipeline}.

The fine-text encoder takes fine-grained motion descriptions with named step marks as input, and outputs step embeddings encoding the full features of each step. In order to better capture the order among steps, we propose a new step-aware self-attention, which adds hard positional embeddings to the step embeddings and then passes them through one self-attention block. The coarse-text encoder digests the coarse-grained description and outputs the corresponding token embeddings. Outputs from the fine-text and coarse-text encoder are then put into the fine-coarse cross-attention block. For the diffusion block, We utilize the diffusion model architecture from \citet{MotionDiffuse} to fuse textual and motion features, and conduct training and inference.

\paragraph{Fine-text Encoder.} We manage to overcome the input length limit of CLIP by respectively encoding each step. In the fine-text encoder, we first extract the text of each step from fine-grained descriptions ($\textbf{STEPk}$), pass it through CLIP, and acquire the step embedding ($step\_embs_{[k]}$) by averaging all token representations.
\begin{equation}
    \label{step_embedding}
    step\_embs_{[k]} = Ave(CLIP(\textbf{STEPk}))
\end{equation}

\paragraph{Step-aware Self-Attention.} Given step embeddings, our step-aware self-attention mechanism adds hard positional embeddings ($PE$) to them through sine and cosine functions from \citet{attention}, and feeds these position-encoded step embeddings into a self-attention Transformer block ($SelfAtt$) to get fine-grained textual features ($f\_feas$). This mechanism helps better capture the temporal relationships among steps, and improve the quality of fine-text features through interactions among step embeddings.
\begin{equation}
\begin{split}
    \label{fine_text_features}
    f\_feas = SelfAtt(PE(step\_embs))
\end{split}
\end{equation}

\paragraph{Coarse-text Encoder.} A coarse-grained description summarizes key points of the whole motion, so we also intend to use such high-level instruction-like information. We pass it through CLIP, and acquire coarse-grained textual features with starting and ending token embeddings.

\paragraph{Fine-coarse Cross-Attention.} Now we get both fine and coarse textual features. Coarse features encode short instructions summarizing key points of the whole motion sequence, while fine features encode detailed step-based descriptions along with explicit time information for each step. We then feed the two types of features into our fine-coarse cross-attention Transformer block to align and combine their information, followed by an FFN to get the final text embeddings.

\paragraph{Diffusion Block.} The diffusion process is composed of a sequence of diffusion steps, each of which is in need of a diffusion step embedding to combine with the diffusion sequence number $t$. Therefore, we directly use the embedding corresponding to token [E] of the coarse text in text embeddings, as shown in Figure \ref{pipeline}. Equipped with the text embeddings and diffusion step embedding, we simply feed them into the diffusion model block, and conduct training and inference similar to MotionDiffuse \citep{MotionDiffuse}.

\section{Experiments}

We design our experiments to answer the following questions: 1) What are the benefits of different blocks and mechanisms in FineMotionDiffuse? 2) What advantages does FineMotionDiffuse (trained on FineHumanML3D) own over MotionDiffuse (trained on HumanML3D)? 3) Can FineMotionDiffuse deal with spatially or chronologically composite motions?

\subsection{Settings}

\paragraph{Datasets.} We use the 85,646 fine-grained motion descriptions along with the 29,228 motions from our FineHumanML3D dataset. For coarse-grained motion descriptions, we use the counterparts of our fine-grained descriptions in the original HumanML3D dataset, counting to 85,646. The training dataset consists of 23,966 motions with 70,222 descriptions, while the testing dataset consists of 5,262 motions with 15,424 descriptions.

\begin{table*}[htbp]
\centering
\begin{tabular}{ccccc}
\toprule[1pt]
\multirow{2}{*}{\textbf{Methods}} & \multirow{2}{*}{\textbf{FID$\downarrow$}} & \multicolumn{2}{c}{\textbf{R-precision$\uparrow$}} & \multirow{2}{*}{\textbf{Diversity$\rightarrow$}}\\
 & & \textbf{Top 1} & \textbf{Top 2} & \\
\midrule[1pt]
Real motions & $0.002^{\pm.000}$ & $0.469^{\pm.002}$ & $0.666^{\pm.002}$ & $9.819^{\pm.055}$\\
\midrule
MotionDiffuse\_AddFC & $1.768^{\pm.031}$ & $0.383^{\pm.002}$ & $0.572^{\pm.002}$ & $9.447^{\pm.115}$\\
FineMotionDiffuse & $1.389^{\pm.020}$ & $0.379^{\pm.002}$ & $0.567^{\pm.002}$ & $9.608^{\pm.107}$\\
\bottomrule[1pt]
\end{tabular}
\caption{\label{main_results}
\textbf{Quantitative evaluation results.} While $\uparrow$ and $\downarrow$ respectively mean the higher the better and the lower the better, $\rightarrow$ means results are better when closer to those of the real motions.
For a fair comparison, each evaluation is run 20 times with $\pm$ denoting the 95\% confidence interval.
}
\end{table*}

\paragraph{Evaluation Metrics.} Following~\citet{Guo2022GeneratingDiverseA}, we pre-train a text-motion contrastive model on the training set of FineHumanML3D for evaluation purposes. With the text encoder and motion encoder from this model, we apply three evaluation metrics. 1) \textit{Frechet Inception Distance (FID)}: The generated motion and its ground-truth motion are passed through the motion encoder, and FID is the distance between the two encoded motion sequences. 2) \textit{R-precision}: For one text and its generated motion sequence, 31 mismatched texts are sampled from the testing set. The motion sequence is encoded by the motion encoder, and 32 texts are encoded by the text encoder. The top-k (k=1,2) accuracy is calculated by checking whether the true text falls into the top-k nearest texts from the motion. 3) \textit{Diversity}: All generated motion sequences in the testing set are randomly split into pairs, and the average joint difference is calculated as the diversity metric.

\paragraph{Baseline.} We build on MotionDiffuse~\citep{MotionDiffuse} to propose MotionDiffuse\_AddFC as our baseline model, following an intuitive idea to fuse fine-grained textual features by simply adding tensors of each step embedding. It first respectively encodes the coarse-grained description and each step in the fine-grained description with CLIP, which results in $1+n$ tensors of the size (context\_length, token\_embedding\_dim). Then we directly add the tensors at each token position, resulting in one fine-coarse feature tensor.

\paragraph{Implementation Details.} In our text encoder, CLIP ViT-B/32 \citep{CLIP} is applied to encode both fine-grained and coarse-grained descriptions. The self-attention block and fine-coarse cross-attention block are both composed of four Transformer encoder layers, and the latent dimensions remain 256. We run the main experiments on 8 A100-PCIE-40GB with the total batch size of 2048. \citet{MotionDiffuse} run 100K iterations on MotionDiffuse, which would be expensive in our case. Therefore, the total
number of our iterations is 5.2K. The rest parameter settings remain the same as MotionDiffuse.

\subsection{Main Results}

As shown in Table \ref{main_results}, 
while the two models are comparable in R-precision, FineMotionDiffuse significantly improves FID by 0.38. This indicates the hierarchical attention architecture helps FineMotionDiffuse produce more accurate motion sequences, aligning more precisely with the ground-truth motions. Additionally, FineMotionDiffuse's improvement in Diversity shows its stronger ability to recognize details from fine-grained descriptions.

\subsection{Ablation Study}

For the ablation study, we investigate the following variants of our method: 1) Unfreezing the CLIP parameters; 2) Different choices of step representations; 3) The effect of fine-coarse cross-attention; 4) Truncating input step texts.

\begin{table}[h]
\centering
\begin{tabular}{cc}
\toprule[1pt]
\textbf{Methods} & \textbf{FID$\downarrow$}\\
\midrule[1pt]
FineMotionDiffuse & $1.645^{\pm.084}$\\
\midrule
$+$ CLIP\_unfrozen & $1.777^{\pm.067}$\\
ave$\rightarrow$[E] & $2.140^{\pm.080}$\\
$-$ cross-attention & $2.839^{\pm.129}$\\
\midrule
delFirstLast & $2.193^{\pm.093}$\\
delInner & $2.228^{\pm.138}$\\
delFirstLast$_{input}$ & $1.511^{\pm.104}$\\
delInner$_{input}$ & $3.117^{\pm.075}$\\
\bottomrule[1pt]
\end{tabular}
\caption{\label{ablation_results}
\textbf{Quantitative evaluation results of ablation experiments.} Each evaluation is run 10 times with $\pm$ denoting the 95\% confidence interval.
}
\end{table}

Due to high GPU consumption when training and evaluating the whole dataset, we take 2,046 motions for training from the whole training set, and 878 motions for evaluation from the whole testing set. Table \ref{ablation_results} reports quantitative evaluation results of these ablated variants, where FID is taken for comparison.

\paragraph{Model Ablation.} After we unfreeze parameters in CLIP ($+$ CLIP\_unfrozen), FID increases by 0.13. We guess that freezing these parameters preserves the generalization ability of the model.

Next, we take the embedding of [E] token of each step (ave$\rightarrow$[E]), instead of averaging along all token embeddings. FID increases by 0.5, showing the advantage of acquiring a step embedding by taking the average. Further, we remove the cross-attention with coarse-grained descriptions ($-$ cross-attention), and find that FID increases by a large margin of 1.2. This proves that the cross-attention between fine-grained and coarse-grained descriptions greatly helps the model generate more precise motions.

\paragraph{Input Ablation.} To specify the roles of fine-grained descriptions of each step, we conduct two groups of ablation experiments. For the first group: 1) We train and test using truncated data by removing the first and last steps (referred to as delFirstLast); 2) We train and test using the first and last steps only (referred to as delInner). Their performance drops in FID show that the integrity of fine-grained descriptions in training data is important for the model performance.

For the second group, we directly use our originally trained FineMotionDiffuse, but make some modifications when testing: 1) We remove the first and last steps (referred to as delFirstLast$_{input}$), and it is quite surprising to find that FID improves by 0.13; 2) We keep the first and last steps only when testing (referred to as delInner$_{input}$), and find that it suffers from severe performance drops. Careful investigation into the fine-grained descriptions reveals that the first and last steps usually describe beginning and ending poses, which are mostly motionless states compared with actions in those inner steps, and do not vary much across different motions. Therefore, we think that FineMotionDiffuse trained on FineHumanML3D is more sensitive to rich actions in the inner steps, and its generalization ability tends to be disturbed by the relatively uninformative first and last steps.

\subsection{Qualitative Analysis}
We aim to compare motions generated by FineMotionDiffuse and MotionDiffuse, by investigating: 1) How well they conform to the input descriptions; 2) How well motion details are performed, e.g., whether feet slide on the ground, whether body parts shake frequently, etc. Therefore, we collect 30 coarse-grained motion descriptions (9 for seen basic motions and 21 for unseen composite motions), and expand them into fine-grained ones with the LLM, using our prompt template.

For each motion type, we generate three motion sequences with three methods: 1) Feeding the fine-grained description along with the coarse-grained one into FineMotionDiffuse; 2) Feeding the coarse-grained description into MotionDiffuse (MotionDiffuse\_coarse); 3) Feeding the fine-grained description (named step marks removed) into MotionDiffuse (MotionDiffuse\_detailed), with the texts being truncated due to the 77-token input length limit of CLIP.

We conduct human evaluation on all generated sequences. Each sequence is scored from 1 to 5, with 0.5 as a unit. We respectively calculate the average scores of all motion sequences, basic motion sequences, and composite motion sequences for each method. Additionally, we also report the score drops from basic to composite motions ($\Delta$). Results are shown in Table \ref{qualitative_analysis}.

\begin{table}[htbp]
\scriptsize
\centering
\begin{tabular}{ccccc}
\toprule[1pt]
\textbf{Methods} & \textbf{all$\uparrow$} & \textbf{basic$\uparrow$} & \textbf{composite$\uparrow$} & \textbf{$\Delta\downarrow$}\\
\midrule
FineMotionDiffuse & 3.97 & 4.33 & 3.81 & 0.52\\
MotionDiffuse\_coarse & 3.55 & 3.94 & 3.38 & 0.56\\
MotionDiffuse\_detailed & 2.55 & 3.11 & 2.31 & 0.80\\
\bottomrule[1pt]
\end{tabular}
\caption{\label{qualitative_analysis} Human evaluation on 30 motions.}
\end{table}

Looking into the scores, we find that MotionDiffuse\_detailed performs the worst, displaying its failure in adapting to fine-grained descriptions. To be more concrete, when fed with fine-grained descriptions, MotionDiffuse performs poorly in recognizing appropriate body parts and movements.

Meanwhile, FineMotionDiffuse performs better than MotionDiffuse\_coarse, which proves motion details are more effectively learned during its training. Furthermore, FineMotionDiffuse has the smallest score drop from basic to composite motions, showing a stronger generalization ability to deal with unseen composite motion descriptions.

\subsection{Case Study}

Here we investigate model performance on composite motions from two dimensions, spatial and chronological compositionality. Spatial compositionality denotes the model's ability to simultaneously perform two basic motions, and chronological compositionality indicates performing two basic motions in a sequential manner. We choose two groups of cases for qualitative analysis. Full details of the input descriptions are in Appendix \ref{appendix_cases}.

\begin{figure}[htbp]
	\centering
	\subfigbottomskip=2pt
	\subfigcapskip=-5pt
	\subfigure[FineMotionDiffuse, "A person slightly squats."$\color{red}(C+F)$]{
        \label{spatial_comp:b1}
		\includegraphics[width=\linewidth]{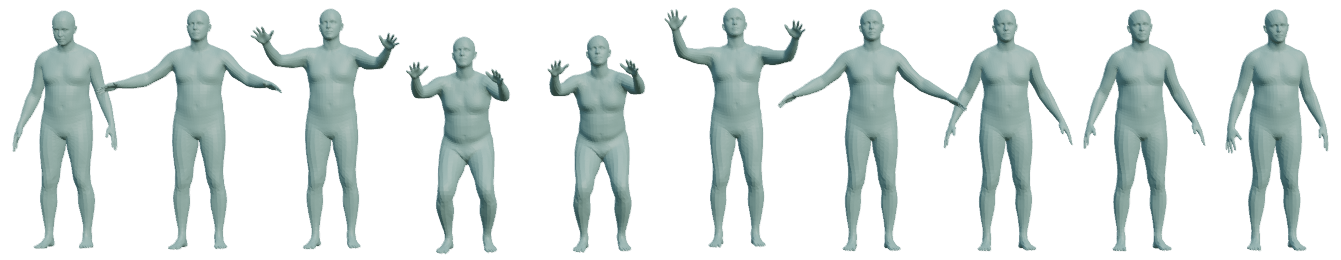}}
        \\
	\subfigure[FineMotionDiffuse, "A man raises both arms above head."$\color{red}(C+F)$]{
        \label{spatial_comp:b2}
		\includegraphics[width=\linewidth]{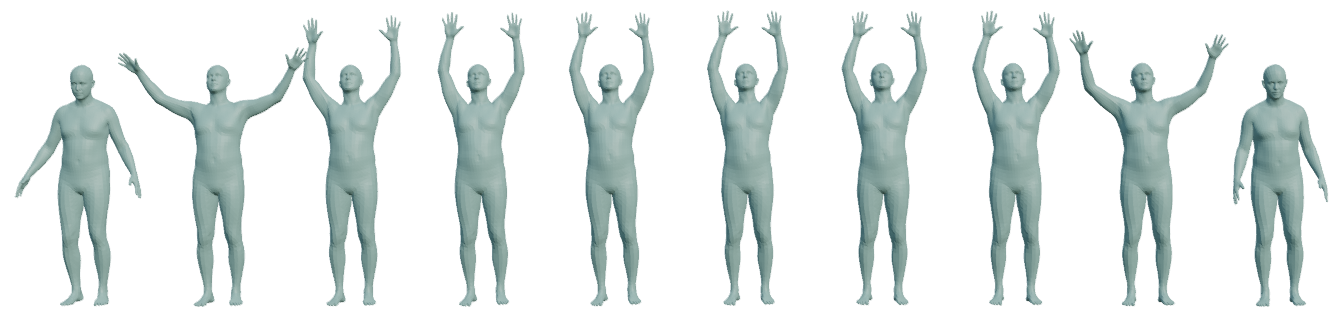}}
        \\
        \subfigure[FineMotionDiffuse, "A man slightly squats with both arms raised above head."$\color{red}(C+F)$]{
        \label{spatial_comp:fm}
		\includegraphics[width=\linewidth]{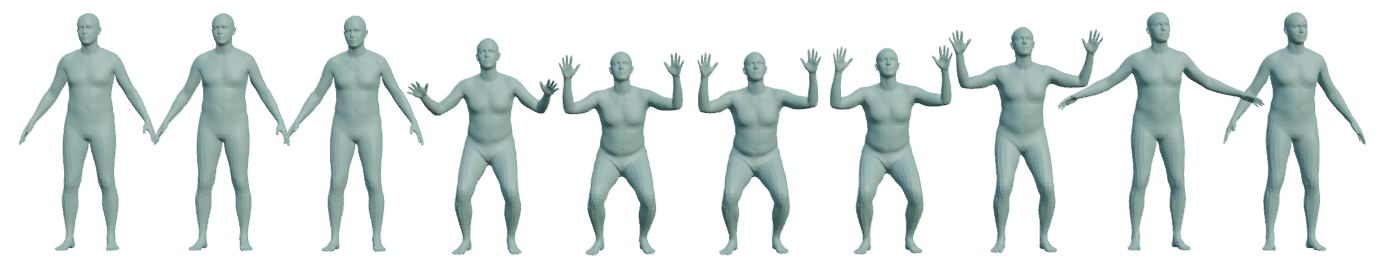}}
        \\
	\subfigure[MotionDiffuse, "A man slightly squats with both arms raised above head."$\color{red}F$]{
        \label{spatial_comp:m_f}
		\includegraphics[width=\linewidth]{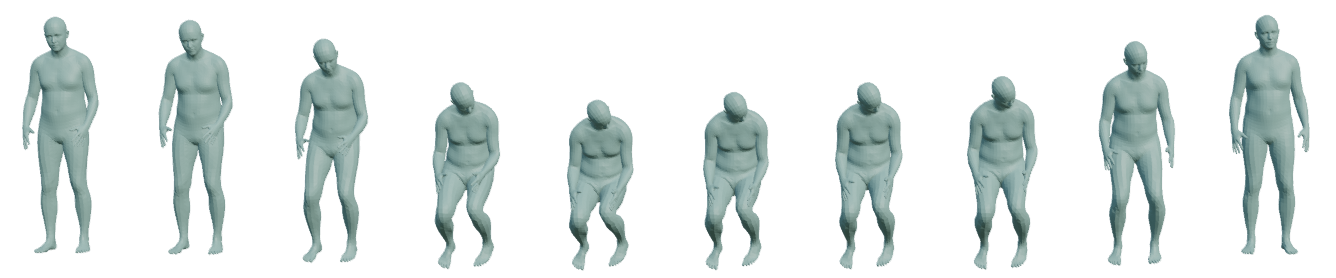}}
	  \\
	\subfigure[MotionDiffuse, "A man slightly squats with both arms raised above head."$\color{red}C$]{
        \label{spatial_comp:m_c}
		\includegraphics[width=\linewidth]{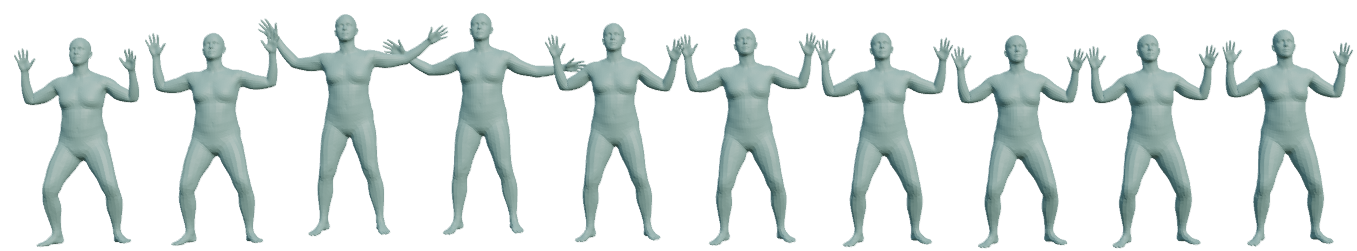}}
	\caption{\textbf{Example motions with spatial compositionality.} $\color{red}C$ denotes inputting coarse-grained texts, $\color{red}F$ denotes inputting fine-grained texts, and $\color{red}C+F$ denotes inputting both texts. Due to space limits, we only display coarse-grained descriptions here. Fine-grained descriptions are shown in Appendix \ref{appendix_cases}.}
\end{figure}

\paragraph{Spatial Compositionality.} While FineMotionDiffuse respectively learns how to slightly squat (Figure \ref{spatial_comp:b1}) and raise both arms above the head (Figure \ref{spatial_comp:b2}), it is able to compose the corresponding composite motion that simultaneously performs the two basic motions (Figure \ref{spatial_comp:fm}). In contrast, MotionDiffuse fails to handle the motions of arms when dealing with fine-grained texts (Figure \ref{spatial_comp:m_f}). Taking the coarse-grained texts as input, MotionDiffuse does not show the process of squatting down in the beginning (Figure \ref{spatial_comp:m_c}).

\begin{figure}[htbp]
	\centering
	\subfigbottomskip=2pt
	\subfigcapskip=-5pt
	\subfigure[FineMotionDiffuse, "A man walks."$\color{red}(C+F)$]{
        \label{chro_comp:b1}
		\includegraphics[width=\linewidth]{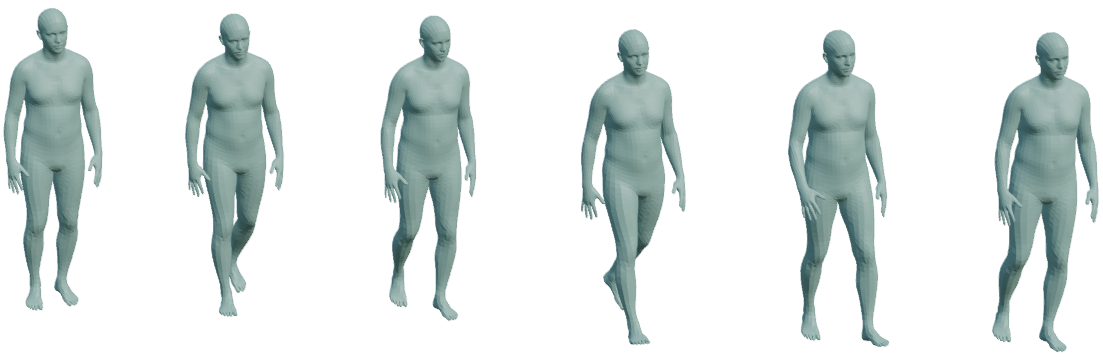}}
        \\
	\subfigure[FineMotionDiffuse, "A man kicks with one leg."$\color{red}(C+F)$]{
        \label{chro_comp:b2}
		\includegraphics[width=\linewidth]{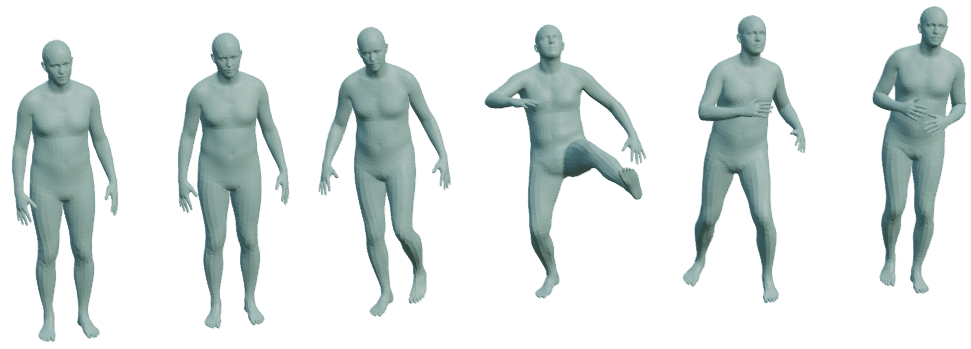}}
        \\
        \subfigure[FineMotionDiffuse, "A man walks, then kicks with one leg."$\color{red}(C+F)$]{
        \label{chro_comp:fm}
		\includegraphics[width=\linewidth]{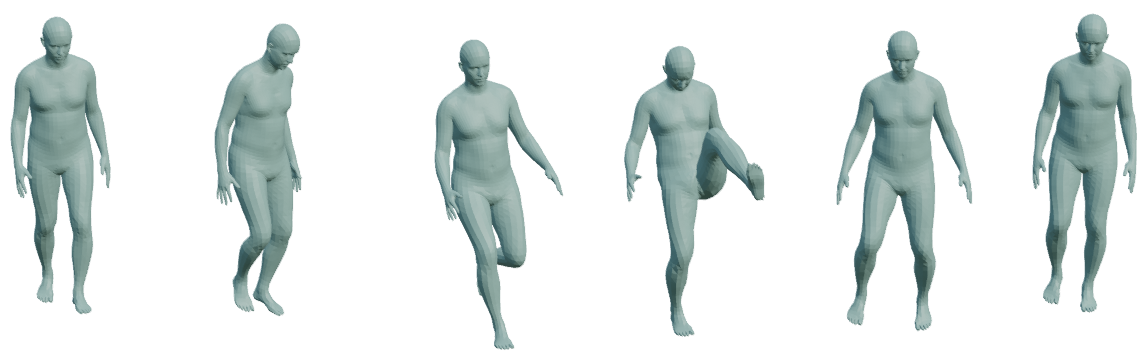}}
        \\
	\subfigure[MotionDiffuse, "A man walks, then kicks with one leg."$\color{red}F$]{
        \label{chro_comp:m_f}
		\includegraphics[width=0.9\linewidth]{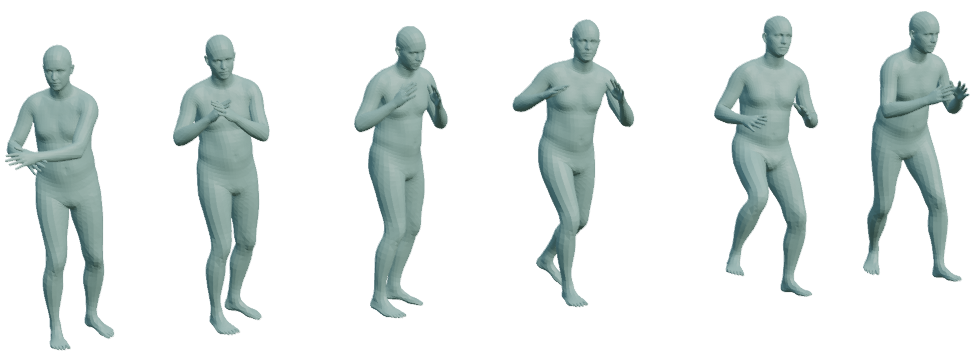}}
        \\
	\subfigure[MotionDiffuse, "A man walks, then kicks with one leg."$\color{red}C$]{
        \label{chro_comp:m_c}
		\includegraphics[width=0.9\linewidth]{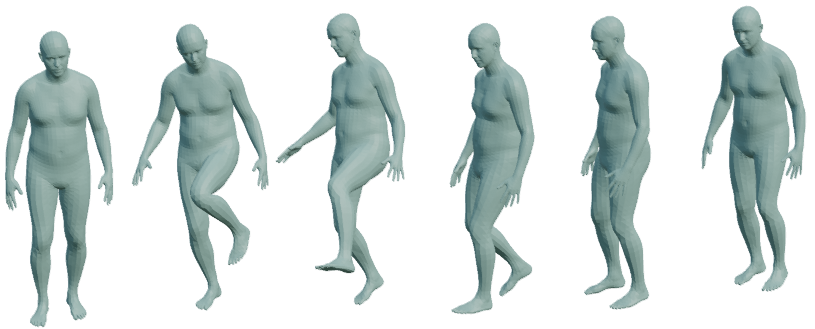}}
	\caption{\textbf{Examples with chronological compositionality.} Fine-grained descriptions are in Appendix \ref{appendix_cases}.}
\end{figure}

\paragraph{Chronological compositionality.} While FineMotionDiffuse respectively learns how to walk (Figure \ref{chro_comp:b1}) and kick with one leg (Figure \ref{chro_comp:b2}), it successfully composes the corresponding composite motion that concatenates the two basic motions in time order (Figure \ref{chro_comp:fm}). In contrast, MotionDiffuse fails to kick when dealing with fine-grained texts (Figure \ref{chro_comp:m_f}). Taking the coarse-grained texts as input, MotionDiffuse fails to make the agent walk~(Figure \ref{chro_comp:m_c}).

\paragraph{Analysis.}In the two groups of cases, FineMotionDiffuse successfully shows the ability of handling spatial and chronological compositionalities, while MotionDiffuse fails. On the one hand, MotionDiffuse trained on large amounts of coarse-grained texts does not learn mappings from fine-grained motion-related descriptions to motion primitives, which explains its failure in spatial and chronological composition when dealing with fine-grained texts. On the other hand, when inputs are coarse-grained composite descriptions, MotionDiffuse cannot show the whole motions. It might only learn how to map a seen coarse text into a single motion, but fails in acquiring the ability to generate an unseen composite motion from the combination of seen coarse texts.

\section{Conclusion}

In this paper, we drive the LLM to build FineHumanML3D, the first large-scale language-motion dataset
specializing in fine-grained textual descriptions,  paving the road to investigate fine-grained text2motion. We find that a step-by-step instruction with a pseudo-code compulsory check can greatly improve the quality of generated descriptions. We propose FineMotionDiffuse to make the best of both fine- and coarse-grained descriptions. Both quantitative and qualitative evaluations show that FineMotionDiffuse trained on FineHumanML3D can better learn the mappings from fine-grained descriptions and exhibit stronger generalizability compared to previous state-of-the-art models, especially when dealing with spatially and chronologically composite motions. In the future, we hope to build on such methods to scale towards increasingly complex human behaviors.

\section*{Limitations}
\noindent\textbf{Dataset.\quad}The fine-grained descriptions in FineHumanML3D come from an LLM pretrained on extensive texts and codes. Although the LLM learns lots of knowledge about human bodies and physical motions, the direct expansion from coarse-grained motion descriptions unavoidably deviates from the golden motion sequences, which, to some degree, leads to the misbehavior of newly trained FineMotionDiffuse. We expect future work to automatically check and correct the generated descriptions, so that they conform to the ground-truth motions. Also, large pre-trained multimodal models like GPT-4 \citep{GPT4} can be used for dataset construction. Even more, the golden motions may be converted to some kinds of LLM-friendly inputs, so that LLM directly receives exact motion information, which helps generate more high-quality fine-grained descriptions.

\noindent\textbf{Model.\quad}The current diffusion-based model architecture is very slow in both training and inference, which hinders its application in real-world scenarios. Therefore, several acceleration methods \citep{EfficientDM1, EfficientDM2} can be adopted to alleviate this problem. Moreover, as some VAE-based models \citep{T2M-GPT} are reported to acquire comparable performance to diffusion-based ones, our method also shows potential in application to such architectures.

\section*{Ethics Statement}
The original dataset we use, HumanML3D, is not published directly, due to the distribution policy of one of its origins, AMASS dataset. Therefore, we follow the official instructions from the builders of HumanML3D (\href{https://github.com/EricGuo5513/HumanML3D}{https://github.com/EricGuo5513/HumanML3D}) to reproduce this dataset. Throughout the work, we only call OpenAI's public API for GPT-3.5-turbo-0301. All experiments along with the dataset construction cost less than \$350. The text tokenizer and POS tagger we use come from an open-source Python package spaCy 3.5 (\href{https://spacy.io/usage/v3-5}{https://spacy.io/usage/v3-5}). We recruit our colleagues to conduct human scoring and fairly pay them with more than \$10 per hour.

\section*{Acknowledgement}
This work is supported by NSFC (62161160339). We would like to thank the anonymous reviewers for their helpful comments and suggestions. We would also like to thank Jason Narowdowsky for helping with the rebuttal and proofreading, and Yusuke Miyao for helping with the rebuttal.

\bibliography{custom}
\bibliographystyle{acl_natbib}

\appendix

\clearpage
\onecolumn
\section{Our Prompts and Evaluation}
\label{appendix_prompts}

{
\small
\begin{longtable}[htbp]{p{3.7cm}p{11cm}}
    \caption{Our prompts} \label{Appendix: prompts} \\
    \toprule[2pt] \textbf{Prompts} & \textbf{Contents}\\ \midrule[2pt]
    \endfirsthead
    {{\bfseries \tablename\ \thetable{} (continued)}} \\
    \toprule[2pt] \textbf{Prompt Name} & \textbf{Contents}\\ \midrule[2pt]
    \endhead
    \multicolumn{2}{l}{{Continued on the next page}} \\ 
    \endfoot
    \bottomrule[2pt]
    \endlastfoot

    \textit{P1} & $\textbf{[system]}$ Please provide a detailed expansion of the following action description. The new description should be in chronological order. It should specify spatial position changes (including angle changes) of relevant body parts and muscle tension.\newline$\textbf{[user]}$ \{THE USER TEXT\}\\
    \midrule
    \textit{P2} &
    $\textbf{[system]}$ Please provide a detailed expansion of one coarse-grained motion description. The new description should be in chronological order. It should specify spatial position changes (including angle changes) of relevant body parts.\newline
    $\textbf{[example\ user]}$ A person squats down then jumps.\newline
    $\textbf{[example\ assistant]}$ The person begins in a standing position with his feet shoulder-width apart and his arms at his sides. He initiates the squat by bending his knees and lowering his hips towards the ground. As he descends, his torso remains upright and his arms swing back behind them for balance. To initiate the jump, the person pushes through his feet and extends his legs and hips explosively. As he leaves the ground, his arms swing forward and upward to help propel them upward. His legs remain relaxed in the air. He lands on the balls of his feet and then gradually lowers his heels to the ground, absorbing the impact with a slight bend in his knees.\newline
    $\textbf{[user]}$ \{THE USER TEXT\}\\
    \midrule
    \textit{P3} &
    $\textbf{[system]}$ Please provide a detailed expansion of the following action description. The new description should be in chronological order. It should specify spatial position changes (including angle changes) of relevant body parts. Here is one example. You should first generate thinking, then the paraphrased description.\newline
    $\textbf{[example\ user]}$ A person squats down then jumps.\newline
    $\textbf{[example\ assistant]}$ Thinking: This is an action without tools. It consecutively uses “squat” and “jump”. To expand it, detailed explanations of the two actions and how the person converts from “squat” to “jump” should be specified.\newline
    The paraphrased description: The person begins in a standing position with his feet shoulder-width apart and his arms at his sides. He initiates the squat by bending his knees and lowering his hips towards the ground. As he descends, his torso remains upright and his arms swing back behind them for balance. As he reaches the bottom of the squat, his thighs remain parallel to the ground. To initiate the jump, the person pushes through his feet and extends his legs and hips explosively. As he leaves the ground, his arms swing forward and upward to help propel them upward. His legs remain relaxed in the air. He lands on the balls of his feet and then gradually lowers his heels to the ground, absorbing the impact with a slight bend in his knees.\newline
    $\textbf{[user]}$ \{THE USER TEXT\}\\
    \midrule
    \textit{P4} &
    $\textbf{[system]}$ Please provide a detailed expansion of one coarse-grained motion description. The new description should be in chronological order and step by step. It should specify spatial position changes (including angle changes) of relevant body parts.\newline
    $\textbf{[example\ user]}$ A person squats down then jumps.\newline
    $\textbf{[example\ assistant]}$ <step 1>The person begins in a standing position with his feet shoulder-width apart and his arms at his sides. He initiates the squat by bending his knees and lowering his hips towards the ground. As he descends, his torso remains upright and his arms swing back behind them for balance.</step 1> <step 2>To initiate the jump, the person pushes through his feet and extends his legs and hips explosively. As he leaves the ground, his arms swing forward and upward to help propel them upward. His legs remain relaxed in the air.</step 2> <step 3>He lands on the balls of his feet and then gradually lowers his heels to the ground, absorbing the impact with a slight bend in his knees.</step 3>\newline
    $\textbf{[user]}$ \{THE USER TEXT\}\\
    \midrule
    \textit{P5} &
    $\textbf{[system]}$ Please provide a detailed expansion of one coarse-grained motion description. The new description should be in chronological order and step by step. It should specify spatial position changes (including angle changes) of relevant body parts.\newline
    $\textbf{[example\ user]}$ A person squats down then jumps.\newline
    $\textbf{[example\ assistant]}$ <step 1: beginning pose>The person begins in a standing position with his feet shoulder-width apart and his arms at his sides.</step 1: beginning pose> <step 2: squat down>He initiates the squat by bending his knees and lowering his hips towards the ground. As he descends, his torso remains upright and his arms swing back behind them for balance.</step 2: squat down> <step 3: jump>To initiate the jump, the person pushes through his feet and extends his legs and hips explosively. As he leaves the ground, his arms swing forward and upward to help propel them upward. His legs remain relaxed in the air.</step 3: jump> <step 4: ending pose>He lands on the balls of his feet and then gradually lowers his heels to the ground, absorbing the impact with a slight bend in his knees.</step 4: ending pose>\newline
    $\textbf{[user]}$ \{THE USER TEXT\}\\
    \midrule
    \textit{P6} &
    $\textbf{[system]}$ In the first paragraph, please provide a detailed expansion of one coarse-grained motion description. The new description should be in chronological order and step by step. It should specify spatial position changes (including angle changes) of relevant body parts. It should not specify any information related to muscles. In the second paragraph, please convert the description in the first paragraph into a pseudo-code format.\newline
    $\textbf{[example\ user]}$ A person squats down then jumps.\newline
    $\textbf{[example\ assistant]}$ <step 1: beginning pose>The person begins in a standing position with his feet shoulder-width apart and his arms at his sides.</step 1: beginning pose> <step 2: squat down>He initiates the squat by bending his knees and lowering his hips towards the ground. As he descends, his torso remains upright and his arms swing back behind them for balance.</step 2: squat down> <step 3: jump>To initiate the jump, the person pushes through his feet and extends his legs and hips explosively. As he leaves the ground, his arms swing forward and upward to help propel them upward. His legs remain relaxed in the air.</step 3: jump> <step 4: ending pose>He lands on the balls of his feet and then gradually lowers his heels to the ground, absorbing the impact with a slight bend in his knees.</step 4: ending pose>\newline
    <step 1: beginning pose>stand(feet shoulder-width apart, arms at sides)</step 1: beginning pose> <step 2: squat down>squat(bending knees, lowering hips towards the ground, torso remaining upright, arms swinging back for balance)</step 2: squat down> <step 3: jump>jump(pushing through feet, extending legs and hips explosively, arms swinging forward and upward, legs remaining relaxed in the air)</step 3: jump> <step 4: ending pose>land(landing on the balls of his feet, lowering heels to the ground, slight bending in knees.)</step 4: ending pose>\newline
    $\textbf{[user]}$ \{THE USER TEXT\}\\
    \midrule
    \textit{P7} &
    $\textbf{[system]}$ In the first paragraph, please provide a detailed expansion of one coarse-grained motion description. The new description should be in chronological order and step by step. It should specify spatial position changes (including angle changes) of relevant body parts. It should not specify any information related to muscles. In the second paragraph, please convert the description in the first paragraph into a pseudo-code format.\newline
    $\textbf{[example\ user]}$ A person squats down then jumps.\newline
    $\textbf{[example\ assistant]}$ <step 1: beginning pose>The person begins in a standing position with his feet shoulder-width apart and his arms at his sides.</step 1: beginning pose> <step 2: squat down>He initiates the squat by bending his knees and lowering his hips towards the ground. As he descends, his torso remains upright and his arms swing back behind them for balance.</step 2: squat down> <step 3: jump>To initiate the jump, the person pushes through his feet and extends his legs and hips explosively. As he leaves the ground, his arms swing forward and upward to help propel them upward. His legs remain relaxed in the air.</step 3: jump> <step 4: ending pose>He lands on the balls of his feet and then gradually lowers his heels to the ground, absorbing the impact with a slight bend in his knees.</step 4: ending pose>\newline
    <step 1: beginning pose>stand(feet shoulder-width apart, arms at sides)</step 1: beginning pose> <step 2: squat down>squat(bending knees, lowering hips towards the ground, torso remaining upright, arms swinging back for balance)</step 2: squat down> <step 3: jump>jump(pushing through feet, extending legs and hips explosively, arms swinging forward and upward, legs remaining relaxed in the air)</step 3: jump> <step 4: ending pose>land(landing on the balls of his feet, lowering heels to the ground, slight bending in knees.)</step 4: ending pose>\newline
    $\textbf{[example\ user]}$ A person standing with his arms out leans backwards into a bridge styled pose.\newline
    $\textbf{[example\ assistant]}$ <step 1: beginning pose>The person begins standing upright with his feet shoulder-width apart, arms extended straight out to sides.</step 1: beginning pose> <step 2: lean backwards>The person shifts his weight onto his heels and slowly leans his entire body backwards, allowing his arms to fall backward.</step 2: lean backwards> <step 3: bridge pose>As his arms touch the ground, he lifts his hips upward, creating a "bridge" with his body. His legs remain straight and his feet remain flat on the ground.</step 3: bridge pose> <step 4: ending pose>The person holds the bridge position for a few seconds before slowly lowering his hips back to the ground and straightening his body to the starting position.</step 4: ending pose>\newline
    <step 1: beginning pose>stand(feet shoulder-width apart, arms extended straight out to sides)</step 1: beginning pose> <step 2: lean backwards>lean\_backwards(shifting weight onto heels, slowly leaning entire body backwards, arms falling backwards)</step 2: lean backwards> <step 3: bridge pose>bridge\_pose(lifting hips upward creating a bridge, legs remaining straight, feet remaining flat on the ground)</step 3: bridge pose> <step 4: ending pose>lower(lowering hips back to the ground, straightening body back to starting position)</step 4: ending pose>\newline
    $\textbf{[user]}$ \{THE USER TEXT\}\\
    \midrule
    \textit{P8} &
    $\textbf{[system]}$ In the first paragraph, please provide a detailed expansion of one coarse-grained motion description. The new description should be in chronological order and step by step. It should specify spatial position changes (including angle changes) of relevant body parts. It should not specify any information related to muscles. In the second paragraph, please convert the description in the first paragraph into a pseudo-code format.\newline
    $\textbf{[example\ user]}$ A person squats down then jumps.\newline
    $\textbf{[example\ assistant]}$ <step 1: beginning pose>The person begins in a standing position with his feet shoulder-width apart and his arms at his sides.</step 1: beginning pose> <step 2: squat down>He initiates the squat by bending his knees and lowering his hips towards the ground. As he descends, his torso remains upright and his arms swing back behind them for balance.</step 2: squat down> <step 3: jump>To initiate the jump, the person pushes through his feet and extends his legs and hips explosively. As he leaves the ground, his arms swing forward and upward to help propel them upward. His legs remain relaxed in the air.</step 3: jump> <step 4: ending pose>He lands on the balls of his feet and then gradually lowers his heels to the ground, absorbing the impact with a slight bend in his knees.</step 4: ending pose>\newline
    <step 1: beginning pose>stand(feet shoulder-width apart, arms at sides)</step 1: beginning pose> <step 2: squat down>squat(bending knees, lowering hips towards the ground, torso remaining upright, arms swinging back for balance)</step 2: squat down> <step 3: jump>jump(pushing through feet, extending legs and hips explosively, arms swinging forward and upward, legs remaining relaxed in the air)</step 3: jump> <step 4: ending pose>land(landing on the balls of his feet, lowering heels to the ground, slight bending in knees.)</step 4: ending pose>\newline
    $\textbf{[example\ user]}$ A person is moving and making gestures as if in a boxing ring.\newline
    $\textbf{[example\ assistant]}$ <step 1: beginning pose>The person begins in a standing position with his feet shoulder-width apart and his arms at his sides.</step 1: beginning pose> <step 2: left punch>He throws a left punch, extending his left arm straight forward while twisting his torso to the left and pivoting his left foot. His right arm remains at his side.</step 2: left punch> <step 3: right punch>He quickly follows with a right punch, extending his right arm straight forward while twisting his torso to the right and pivoting his right foot. His left arm remains at his side.</step 3: right punch> <step 4: left and right hooks>He then throws a left hook, pivoting on his left foot and swinging his left arm in a semi-circular motion, aiming for an imaginary target to his left. He quickly follows with a right hook, pivoting on his right foot and swinging his right arm in a semi-circular motion, aiming for an imaginary target to his right.</step 4: left and right hooks> <step 5: duck>He ducks down, bending his knees and lowering his torso to avoid an imaginary punch.</step 5: duck> <step 6: end pose>He steps back and repeats the sequence, alternatively throwing punches and hooks while changing his stance and defensive movements as if in a boxing match.</step 6: end pose>\newline
    <step 1: beginning pose>stand(feet shoulder-width apart, arms at sides)</step 1: beginning pose> <step 2: left punch>left\_punch(throwing left arm straight forward, twisting torso to left, pivoting left foot, right arm remaining at sides)</step 2: left punch> <step 3: right punch>right\_punch(throwing right arm straight forward, twisting torso to right, pivoting right foot, left arm remaining at sides)</step 3: right punch> <step 4: left and right hooks>left\_hook(pivoting on left foot, swinging left arm in a semi-circular motion); right\_hook(pivoting on right foot, swinging right arm in a semi-circular motion)</step 4: left and right hooks> <step 5: duck>ducking(bending knees, lowering torso)</step 5: duck> <step 6: end pose>repeat(steps 2-5)</step 6: end pose>\newline
    $\textbf{[user]}$ \{THE USER TEXT\}\\
\end{longtable}
}

\begin{table}[htp]
\centering
\begin{tabularx}{\textwidth}{lXXX}
\toprule[2pt]
\textbf{Prompts} & \textbf{Pros} & \textbf{Cons}\\
\midrule[2pt]
\textit{P1} & - & Too many details other than body movements;\newline Too free text formats;\newline Cannot promise time order;\\
\midrule
\textit{P2} & More concise;\newline Mostly in time order; & May exist movements that do not conform to the body constraints;\newline May appear unnecessary statements like the agent's feelings;\\
\midrule
\textit{P3} & Stronger time order; & May sometimes be too detailed (e.g., finger tips);\newline May exist movements that do not conform to the body constraints;\newline May appear unnecessary statements like the agent's imagination;\\
\midrule
\textit{P4} & Much more concise, barely with unnecessary statements;\newline Almost totally in time order;\newline Movements highly conform to body constraints; & Random in splitting steps;\newline May add a useless comment of the whole motion without step marks at the end;\\
\bottomrule[2pt]
\end{tabularx}
\caption{\label{prompts_without_scores}
The pilot prompts we test for text expansion from coarse-grained motion descriptions to fine-grained ones. We test each of them on 10 example coarse-grained descriptions, evaluate fine-grained descriptions returned by GPT-3.5-turbo-0301, and give a qualitative analysis of pros and cons. The pros of each prompt come from comparisons with all prompts above it.
}
\end{table}

\begin{table}[htbp]
\centering
\begin{tabular}{ccc}
\toprule[2pt]
\textbf{Prompts} & \textbf{Averaged Description Level} & \textbf{Averaged Code Level}\\
\midrule[2pt]
\textit{P5} & 1.57 & - \\
\midrule
\textit{P6} & 1.3 & 1.43\\
\textit{P7} & 1.2 & 1.13\\
\textit{P8} & 1.2 & 1.13\\
\bottomrule[2pt]
\end{tabular}
\caption{\label{prompts_with_scores}
The promising prompts we test. Each one of them is tested on 30 coarse-grained texts. For each response from LLM, we score its \textbf{Description Level} (1 for best, 2 for worse, 3 for worst) based on the quality of the returned fine-grained description. We score its \textbf{Code Level} (1 for best, 2 for worse, 3 for worst), except responses from \textit{P5}, based on how well the pseudo-codes conform to the fine-grained description in the response, and whether they omit unnecessary words. For each prompt, its \textbf{Averaged Description Level} is acquired by averaging description levels of all fine-grained descriptions scored by humans, so is the \textbf{Averaged Code Level}. Averaged Description Level denotes the overall quality of returned fine-grained descriptions, and Averaged Code Level denotes the overall quality of returned pseudo codes based on the returned fine-grained descriptions. For both metrics, lower values denote better performance.
}
\end{table}

\twocolumn


\section{Input Texts for Case Study}
\label{appendix_cases}
{\fontsize{9pt}{13.2pt}\selectfont $\color{red}[C]$ for coarse-grained texts; $\color{red}[F]$ for fine-grained texts.}

\subsection*{Spatial Compositionality}

$\color{red}[C]$ A man raises both arms above his head.\\
$\color{red}[F]$ <step 1: beginning pose>The man begins in a standing position with his arms at his sides.</step 1: beginning pose> <step 2: raise arms>He lifts both arms up simultaneously, his hands moving upwards towards the ceiling. His shoulders and elbows straighten as his arms reach their maximum height straight above his head.</step 2: raise arms> <step 3: end pose>He holds this position momentarily before lowering his arms back down to his sides.</step 3: end pose>\\\\
$\color{red}[C]$ A person slightly squats.\\
$\color{red}[F]$ <step 1: beginning pose>The man begins standing upright with his feet hip-width apart and his arms relaxed at his sides.</step 1: beginning pose> <step 2: slight squat>He bends his knees slightly, lowering his hips and shifting his weight slightly towards his heels. His torso remains upright, and his feet remain flat on the ground.</step 2: slight squat> <step 3: end pose>He holds this slightly squatted position.</step 3: end pose>\\\\
$\color{red}[C]$ A man slightly squats with both arms raised above head.\\
$\color{red}[F]$ <step 1: beginning pose>The man begins in a standing position with his feet shoulder-width apart and his arms at his sides.</step 1: beginning pose> <step 2: squat>He slightly squats down, bending his knees and lowering his hips a few inches towards the ground. His torso remains upright and his arms are raised straight above his head, palms facing each other.</step 2: squat> <step 3: end pose>He holds this position for a moment before standing back up again to the starting position with his feet shoulder-width apart and arms at his sides.</step 3: end pose>\\

\subsection*{Chronological Compositionality}

$\color{red}[C]$ A man walks.\\
$\color{red}[F]$ <step 1: beginning pose>The man stands upright with his feet together.</step 1: beginning pose> <step 2: lift foot>He lifts his left foot off the ground and swings it forward while simultaneously shifting his weight onto his right foot.</step 2: lift foot> <step 3: place foot>He plants his left foot on the ground and transfers his weight onto it while simultaneously lifting his right foot off the ground.</step 3: place foot> <step 4: swing foot>He swings his right foot forward while simultaneously shifting his weight onto his left foot, as he prepares to take another step.</step 4: swing foot> <step 5: end pose>He repeats steps 2-4, alternating between his left and right foot, to continue walking.</step 5: end pose>\\\\
$\color{red}[C]$ A man kicks with one leg.\\
$\color{red}[F]$ <step 1: beginning pose>The man begins standing with his feet shoulder-width apart and his arms at his sides.</step 1: beginning pose> <step 2: lift leg>He lifts one leg, bending at the knee and bringing his foot up towards his buttocks.</step 2: lift leg> <step 3: extend leg>He then extends his leg forward, kicking with the foot while keeping the rest of his body stable. His arms remain at his sides for balance.</step 3: extend leg> <step 4: lower leg>After the kick, he lowers his leg back to the ground, returning to his starting position.</step 4: lower leg>\\\\
$\color{red}[C]$  A man walks, then kicks with one leg.\\
$\color{red}[F]$ <step 1: beginning pose>The man begins in a standing position with his feet together and his arms at his sides.</step 1: beginning pose> <step 2: taking a step>He lifts his right foot and takes a step forward with it, placing it on the ground in front of him.</step 2: taking a step> <step 3: kicking>He then swings his left leg forward in a kicking motion, keeping it straight and extending it towards an imaginary target. As he kicks, he leans his torso back slightly for balance.</step 3: kicking> <step 4: end pose>He then lowers his left leg back to the ground and resumes the standing position with his feet together and his arms at his sides.</step 4: end pose>

\end{document}